\documentclass[sigconf,nonacm]{acmart}

\setcopyright{none}
\renewcommand\footnotetextcopyrightpermission[1]{}

\usepackage{algorithm}
\usepackage{algorithmic}
\usepackage{amsmath}
\usepackage{booktabs}
\usepackage{listings}
\graphicspath{{figs/}}
\title{Read Less, Solve More: Token-Efficient Sparse Reading of External Content for AI Agents}
\author{Zedong Liu}
\author{Jiaan Wu}
\author{Xinyang Ma}
\author{Le Xu}
\author{Kai Wang}
\author{Yuanchao Hu}
\author{Dingwen Tao}
\author{Guangming Tan}

\renewcommand{\shortauthors}{Zedong Liu et al.}

\makeatletter
\renewcommand{\@mkauthors@iii}{%
  \global\setbox\mktitle@bx=\vbox{%
    \hsize=\textwidth
    \unvbox\mktitle@bx
    \par\medskip
    \centering
    \parindent=0pt

    {\normalfont\fontsize{12}{15}\selectfont\bfseries
      Zedong Liu\textsuperscript{\mdseries 1,2},
      Jiaan Wu\textsuperscript{\mdseries 1,2},
      Xinyang Ma\textsuperscript{\mdseries 1,2},
      Le Xu\textsuperscript{\mdseries 1,2},\\
      Kai Wang\textsuperscript{\mdseries 3},
      Yuanchao Hu\textsuperscript{\mdseries 3},
      Dingwen Tao\textsuperscript{\mdseries 1,2},
      Guangming Tan\textsuperscript{\mdseries 1,2}
      \par
    }

    \vskip 4pt

    {\normalfont\fontsize{10}{12}\selectfont
      \textsuperscript{1}University of Chinese Academy of Sciences, Beijing, China\\
      \textsuperscript{2}Institute of Computing Technology, Chinese Academy of Sciences, Beijing, China\\
      \textsuperscript{3}Songshan Lake Materials Laboratory, Dongguan, China
      \par
    }

    \bigskip
  }%
}
\makeatother

\begin{document}

\begin{abstract}
Web-facing AI agents repeatedly access webpages, documents, repositories, and tool responses, yet reading interfaces often expose entire objects when only sparse evidence is needed. This over-reading increases token and latency costs and can dilute relevant evidence. We present SparseRead, a reusable reading control layer that governs content admission without modifying the agent model. Its \emph{Read Intent Protocol} tracks evolving needs and source coverage; \emph{Adaptive Reader Backends} select how evidence is located and exposed; and the \emph{Budgeted Read Gate} bounds cumulative raw exposure per episode with faithful full-view fallback. A unified Reader interface supports interchangeable evidence acquisition across heterogeneous resources. Across six models, including Claude Opus 5, and five scenarios, SparseRead reduces total tokens by up to 92.9\% and wall time by up to 89.0\%, preserving or improving task quality in 97\% settings. On Web deep-search tasks, it reduces tokens by 31.0--43.0\% and task completion time by 23.0--37.0\%. Ablations validate the three designs, while integration with three agent frameworks demonstrates portability.
\end{abstract}

\ccsdesc[500]{Computing methodologies~Intelligent agents}
\keywords{LLM agents, sparse reading, reading control, evidence acquisition, token efficiency}

\maketitle

\section{Introduction}
\label{sec:introduction}

Large language model (LLM) agents have extended the scope of language models from single-turn generation to long-horizon task execution~\cite{yao2023react,liu2024agentbench}. By coupling reasoning with tool use and access to external artifacts, agents support increasingly complex applications across software engineering, scientific analysis, and interactive environments~\cite{yang2024sweagent,boiko2023coscientist,zhou2024webarena}. For Web agents, these artifacts include webpages, online reports, code repositories, and structured responses from Web APIs. This capability, however, introduces a markedly different inference profile: each step adds new observations that may remain in subsequent model calls, causing token consumption, latency, and serving cost to grow rapidly with trajectory length. This raises a question for agent infrastructure: what should an agent read from a Web resource before its next decision?

Reading constitutes a major source of this growth. Prior work reports that external observations account for 60--80\% of the tokens processed in agent trajectories~\cite{wang2026swepruner}. Yet an agent may receive an entire page, file, or tool result when its next decision requires only a few source passages. Paired executions further show that unmodified agents can consume up to 17.3\(\times\) more tokens without achieving higher task quality. Excess content is not necessarily benign; by diluting relevant evidence, it can induce erroneous tool use, source confusion, and failures to stop. We refer to this mismatch as over-reading: external artifacts are accessed at object granularity, although each decision often requires only a bounded evidence slice. When admitted early, unnecessary content may also be carried into many later model calls.

\begin{figure}[t]
  \centering
  \includegraphics[width=0.96\columnwidth]{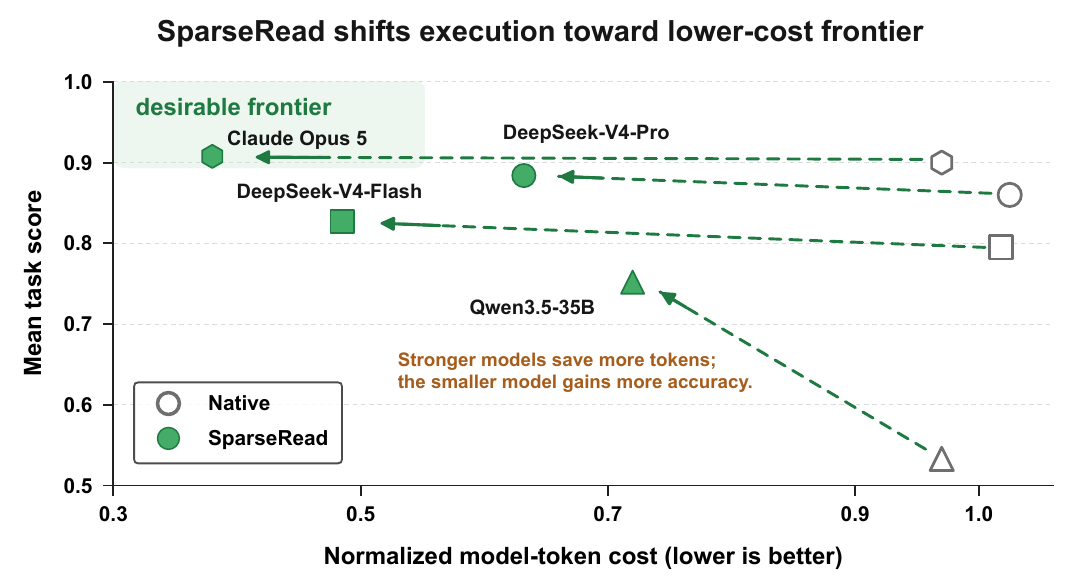}
  \caption{\textbf{Result teaser.} SparseRead
  shifts model execution toward a lower-cost, quality-preserving frontier.}
  \label{fig:intro-pareto}
\Description{Quality versus token and execution-time cost for native reading and SparseRead.}
\end{figure}

Existing techniques address parts of this problem without jointly controlling artifact reading. Long-context architectures and KV-cache compression reduce the compute and memory overhead of context, but do not decide which source content the model should see~\cite{zaheer2020bigbird,NEURIPS2024_28ab4182}. ACON and observation masking reduce tool observations or their persistence in the trajectory~\cite{kang2026acon,lindenbauer2025complexity}. Retrieval, including iterative and agentic approaches, can acquire passages on demand~\cite{asai2024self,xu2024recomp,pmlr-v119-guu20a}, but does not by itself provide a common artifact-level contract for evolving read intents, explicit source coverage, bounded exposure, and full-view fallback. A general mechanism is still needed to govern artifact access before unnecessary content enters the agent context.

Human reading provides a natural abstraction for this missing capability. Human readers first clarify their purpose, inspect an artifact's structure when needed, focus on relevant passages, and revisit unresolved details. We call this capability Sparse Reading: the agent reads only what its current reasoning requires and expands the read only when additional evidence is needed.

Turning this pattern into an agent mechanism raises two challenges. First, information needs change as an agent works through subgoals and verification steps. Each partial read must clarify its purpose, record what has been inspected and what remains uncovered, and use an appropriate way to acquire evidence from the current artifact. Otherwise, the agent may repeatedly revisit the same source. Second, sparse access is not always economical. Small objects and requests that require a complete model-visible view may make repeated partial reads more costly than direct access. These challenges call for explicit evidence state, flexible evidence acquisition, and online control over when to continue reading or fall back to a full view.

\begin{figure}[t]
  \centering
    \includegraphics[width=0.98\columnwidth]{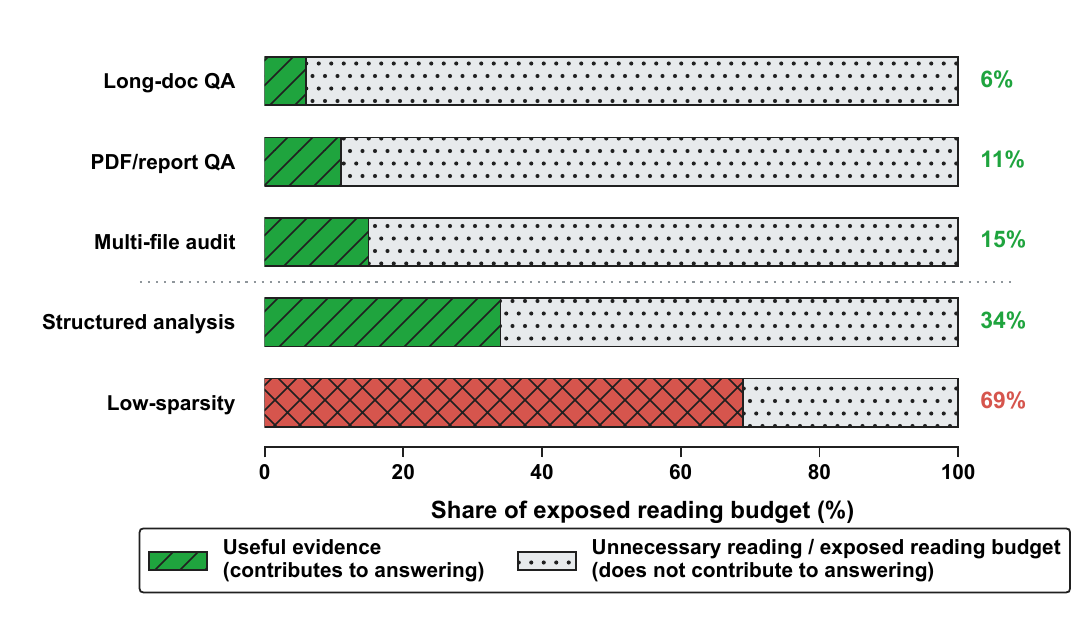}
  \caption{Reading-budget breakdown across representative workloads. Evidence share is the ratio of manually annotated supporting evidence to the source content exposed to the native agent, both measured in characters.}
  \label{fig:overreading-audit}
\Description{Breakdown of useful evidence and over-read content in representative agent workloads.}
\end{figure}

We present \textsc{SparseRead}, a framework that controls artifact admission before broad content enters the model context. The agent states its per-access needs through a Read Intent, and the reading layer returns an EvidencePack containing bounded, source-anchored evidence and explicit coverage; the agent decides whether the evidence is sufficient. SparseRead combines a Read Intent Protocol, Adaptive Reader Backends, and a Budgeted Read Gate. SparseRead can be integrated into existing agent frameworks and used with existing LLMs, without changes to model parameters or a separate generative compression model.

This paper makes the following contributions:

\begin{itemize}
  \item \textbf{Read Intent Protocol.} We introduce a cooperative protocol that turns broad artifact access into a bounded and stateful reading process. It makes per-access evidence needs and source coverage explicit, while leaving sufficiency and stopping decisions to the agent.
  \item \textbf{Adaptive Reader Backends.} We separate reading control from evidence acquisition through a unified Reader interface. Reader configurations combine artifact-specific location or processing methods with exposure forms, selected using calibrated evidence reliability and estimated reading costs.
  \item \textbf{Budgeted Read Gate.} We introduce an inference-time controller that regulates when sparse reading is attempted and when a full view is preferable. An episode-level exposure budget bounds cumulative raw artifact exposure while retaining an explicit full-view fallback.
  \item \textbf{Evaluation.} We compare SparseRead with strong native agents and static and agentic RAG across five scenarios, and examine the contribution and effective regime of each component.
\end{itemize}

\section{Quantifying the Cost of Over-Reading}
\label{sec:over-reading}

We first quantify the sparsity of annotated supporting evidence, then examine how broad exposure accumulates across later calls. Together, these observations make over-reading a measurable agent execution problem.

\begin{figure}[t]
  \centering
  \includegraphics[width=0.98\columnwidth]{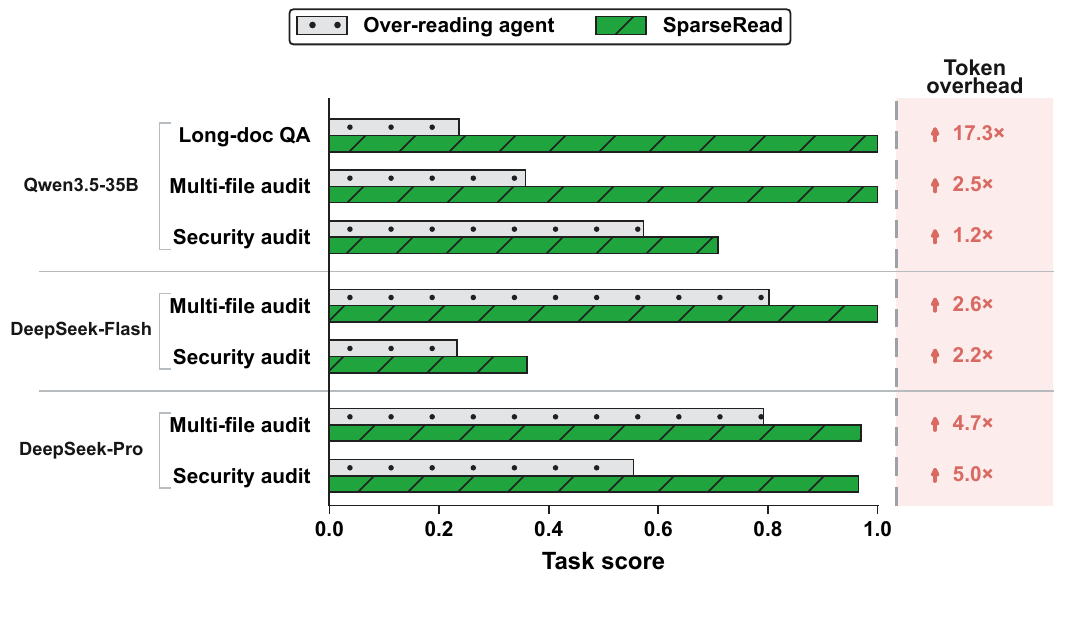}
  \caption{Quality-efficiency comparison. Sparse reading matches or improves task quality while using fewer tokens than full reading. More exposed context does not automatically produce better answers.}
  \label{fig:quality-efficiency}
\Description{Comparison of task quality and token use for sparse and full reading.}
\end{figure}

\subsection{Useful Evidence Is Sparse}
\label{sec:evidence-density}

Figure~\ref{fig:overreading-audit} shows reading-budget breakdowns across representative workloads. For long-document QA, annotated supporting evidence covers a median of 3.17\% of source tokens per question and 2.89\% across five questions from one document in LooGLE; the share is about 5.8\% for analyzed answerable Qasper questions with evidence annotations. These annotation ratios differ from the figure's exposure shares and do not measure the reading needed to locate and verify evidence. Moreover, different questions may require evidence from different parts of the same document. This motivates evidence acquisition guided by the current information need rather than a fixed summary of the entire source.

This observation also highlights three distinct quantities: processing scope, model exposure, and evidence coverage. Processing scope describes how much source content a Reader examines or computes over, whereas exposure measures what enters the model context. A Reader can scan an entire table to compute a column sum while exposing only the result, row count, and source. Coverage, in turn, records which slots in the current Read Intent have supporting evidence and which remain unresolved, without judging sufficiency. These quantities need not scale together. In particular, reducing model exposure does not require reducing the amount of source content processed, allowing Sparse Reading to benefit both localized retrieval and exact whole-object operations.

\begin{figure*}[t]
  \centering
  \includegraphics[width=\textwidth]{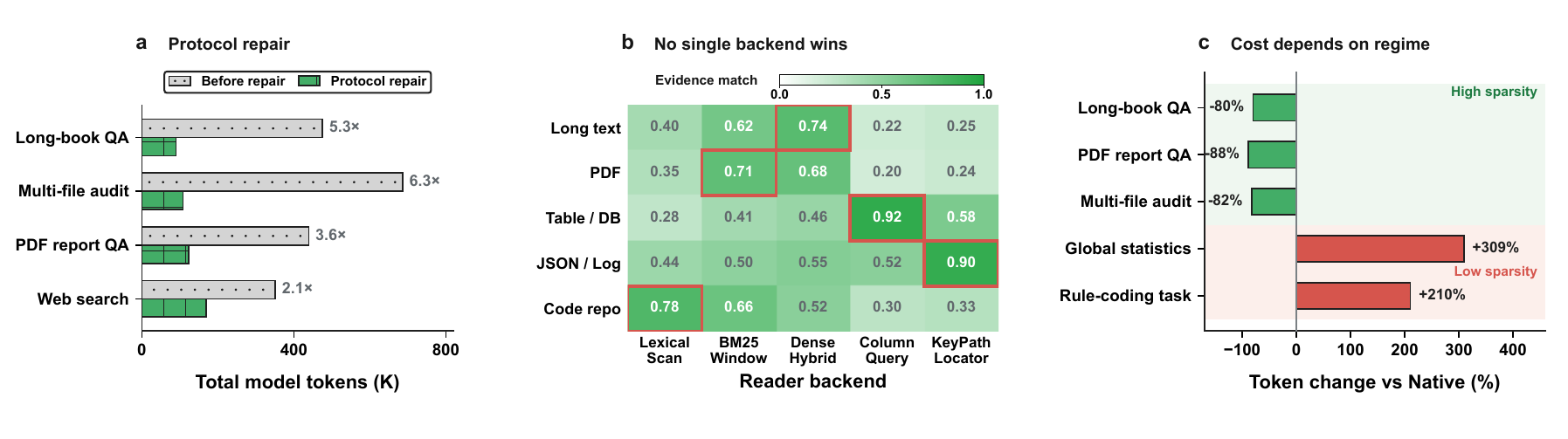}
  \caption{Challenges of controlled reading.
  (a) Protocol repair reduces total model tokens.
  (b) The best evidence location method varies across artifact types.
  (c) Token cost depends on the reading regime.}
  \label{fig:protocol-sufficiency}
  \label{fig:sparsity-regime}
  \Description{Three panels showing protocol-repair costs, evidence matching across Reader backends and artifact types, and token changes across sparse-fit and full-view workloads.}
\end{figure*}

\subsection{Over-Reading Compounds Across Model Calls}
\label{sec:exposure-replay}

The cost of a broad read can extend beyond its initial admission. Content exposed at step \(i\) may remain in later model calls. Let \(e_i\) denote newly exposed content and \(R_i\) the number of subsequent calls retaining it. The resulting replay-weighted exposure is \(C_{\mathrm{exposure}}=\sum_i |e_i|R_i\). This measures logical context replay, not billed tokens or physical prefill work; prefix caching can lower their marginal costs without removing retained content.

If broad reading always improved quality, over-reading would be a cost-only problem. Figure~\ref{fig:quality-efficiency} shows otherwise. In paired high-sparsity runs, full-read agents spend up to 17.3 times more tokens than SparseRead while failing to improve task score. Across model and task groups, selective reading preserves or improves score while reducing exposed context. The result does not imply that less context is always better; it shows that broad context is not a free substitute for reading control.

These observations show why reading should be controlled, but not what evidence an agent should seek, how its coverage is tracked, or when to switch to a full view.

\section{Why Retrieval or Compression Alone Is Not Enough}
\label{sec:reading-challenges}

\textbf{Problem formulation.} At step \(t\), let \(o\) denote the current external artifact and \(h_t\) the current Read Intent. The core challenge is to choose the content \(x_t\) exposed to the model before its next decision, while preserving task-relevant evidence, tracking coverage of \(h_t\), and controlling the cost of subsequent reads.

Retrieval can locate passages, including through iterative queries, while compression can reduce the content admitted to context. Yet neither alone establishes a common reading contract across heterogeneous artifacts. Two challenges remain: tracking evolving evidence needs and deciding when selective reading is no longer economical.

\subsection{Challenge 1: Evolving Needs and Uncertain Evidence Coverage}
\label{sec:evolving-needs}

An agent's information needs evolve during execution. The opening request establishes a session-level objective, but subsequent steps introduce subgoals, follow-up questions, and verification needs. Each access therefore requires an intent specific to the current decision. Iterative retrieval can also update queries; the missing abstraction is a consistent intent and evidence record across artifact types and access tools.

Figure~\ref{fig:protocol-sufficiency}(a) shows that protocol repair reduces total model tokens by 2.1--6.3\(\times\) across long-book QA, multi-file audit, PDF report QA, and Web search. Without a record of inspected regions and unresolved evidence slots, an agent may reopen artifacts or continue searching unnecessarily. Coverage records what has been inspected or found, rather than declaring the evidence sufficient. Source-anchored evidence and explicit coverage help the agent decide what to read or verify next, and when to stop.

Evidence acquisition also depends on the artifact. Figure~\ref{fig:protocol-sufficiency}(b) shows that the best evidence location method varies across artifact types: dense hybrid retrieval works best for long text, while column queries and key-path lookup are more effective for tables and JSON/logs, respectively. Locating evidence and deciding how much to expose are separate choices, motivating configurable Readers rather than a single retrieval pipeline.

\subsection{Challenge 2: The Regime Boundary Is Unknown Before Reading}
\label{sec3.2}

Sparse Reading is well-suited for tasks that require extracting localized information from large artifacts. Long-document QA, PDF comprehension, and audit bundles exhibit this characteristic. However, some tasks favor a full view: small objects may fit directly into context, and some requests require full model-visible access. For these tasks, sparse reading may add overhead without reducing token usage. In contrast, an exact whole-table operation can process the full source yet expose only a compact, verifiable result.

Figure~\ref{fig:protocol-sufficiency}(c) illustrates this boundary. Sparse Reading reduces token consumption by 80--88\% on high-sparsity QA and audit tasks, while forced sparse reading increases tokens by 309\% for global statistics and 210\% for a rule-coding task, where repeated selective access outweighs its savings. Sparse Reading should therefore have an explicit fallback rather than being applied uniformly.

This boundary cannot always be determined before reading: the evidence needed for a decision and the cost of subsequent refinement remain uncertain. Object size or type alone is insufficient. Rather than predicting the optimal regime in advance, an online controller can attempt sparse reading while limiting cumulative exposure against the cost of a faithful full view, switching before the next batch exceeds the budget.

\textbf{Design implication.} Together, these challenges show that Sparse Reading needs more than retrieval or compression. The Read Intent Protocol tracks evolving needs and evidence coverage, Adaptive Reader Backends choose how evidence is located and exposed, and the Budgeted Read Gate controls exposure and full-view fallback, forming a unified reading loop.

\section{SparseRead Design}
\label{sec:design}

\begin{figure}[t]
  \centering
  \includegraphics[width=0.98\columnwidth]{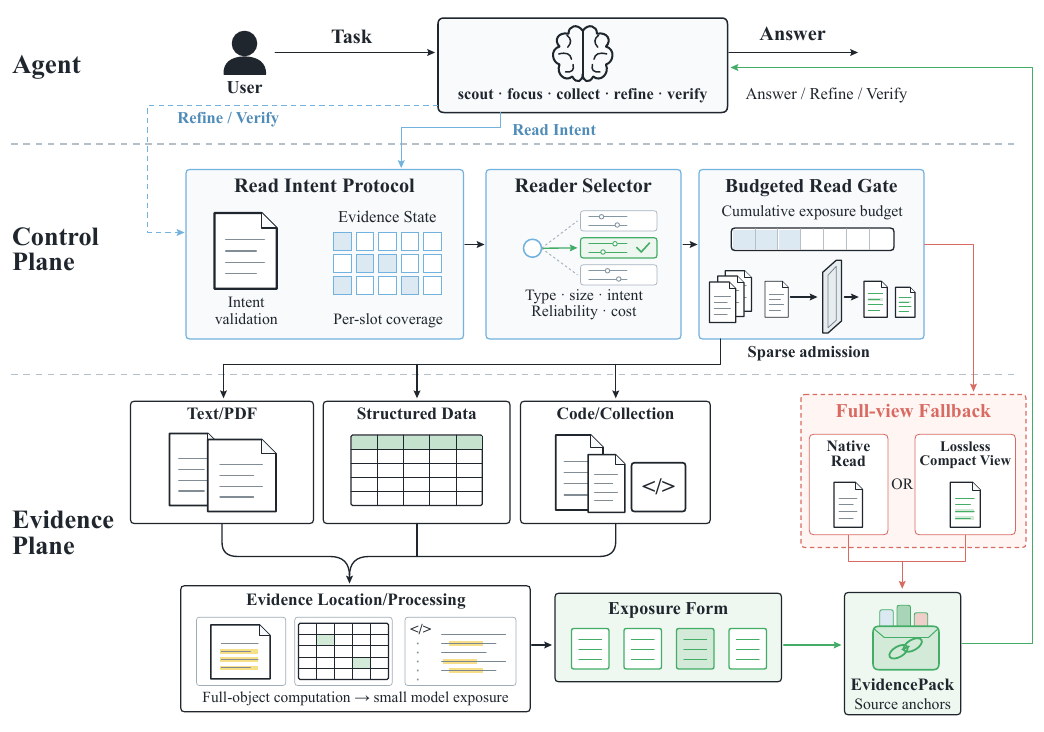}
  \caption{SparseRead separates the agent-facing control plane from replaceable
  Reader Backends. The Read Gate selects the intervention mode, while the
  protocol governs evidence acquisition, state updates, and fallback.}
  \label{fig:framework}
\Description{SparseRead architecture showing the Read Gate, control protocol, and replaceable Reader Backends.}
\end{figure}

SparseRead separates reading control from evidence acquisition. The control plane interprets an agent's reading intent, selects among eligible Reader configurations, maintains the evidence record, and governs what enters the model context. The evidence plane contains Readers that declare their supported artifact types, actions, and execution costs. Readers acquire evidence but do not decide when they are invoked. Figure~\ref{fig:framework} illustrates the architecture. Throughout the design, we distinguish the scope processed by a Reader, the content exposed to the model, and the evidence coverage of the current intent.

The two planes share a common contract: \textbf{Read Intent \(\rightarrow\) EvidencePack}. It covers targeted passages, faithful compact views, and results of exact source operations without changing the agent-facing interface. Ordinary sparse reads leave preceding turns intact, allowing exact-match prefix reuse. A subsequent fallback may retire earlier sparse observations to avoid further replay, potentially losing prefix-cache reuse from the edited point onward.

\subsection{Read Intent Protocol}
\label{sec:agent-protocol}

SparseRead treats reading as a sequence of evidence requests guided by the agent. At each access, the agent supplies a structured \textbf{Read Intent} describing what it needs for its next decision. This access intent takes precedence over a session intent derived from the user's opening request. The session intent fills in unspecified details rather than overriding a new subgoal or follow-up question. An actionable intent proceeds directly to evidence acquisition; an underspecified one receives a bounded \textbf{Object Preview} for refinement. The agent can also request the same preview explicitly through \emph{scout}.

Each read returns an \textbf{EvidencePack} with source-anchored evidence, the scope examined or processed, the amount exposed, and coverage for individual evidence slots. Coverage identifies which needs have supporting evidence and which remain unresolved. It does not establish whether the evidence is sufficient to answer the task. The agent decides whether to answer, refine, or verify; neither the protocol nor the Reader declares completion.

The protocol also maintains the validity and relevance of prior evidence. A SHA-256 fingerprint identifies the artifact version; when it changes, evidence and coverage recorded for that artifact are invalidated. On an unchanged version, earlier spans remain valid but may not cover a new intent. Lightweight validation checks for identifiable slots and consistent scope requirements using structural signals rather than task-specific rules. Vague intents trigger an Object Preview, while inconsistent scope requests are flagged and resolved according to their stated coverage requirements.

The agent interacts through five stable actions, \(\mathcal{A}=\{\mathrm{scout},\allowbreak\mathrm{focus},\allowbreak\mathrm{collect},\allowbreak\mathrm{refine},\allowbreak\mathrm{verify}\}\). The selected Reader maps each supported action to operations over its artifact type, so adding Readers does not expand the agent's action set. A \textbf{reading episode} is a sequence of accesses to one artifact version under a single reading objective. Refinement and verification within an episode share its evidence record and exposure budget, even as the access intent changes. A distinct objective starts a new episode. Algorithm~\ref{alg:sparseread-protocol} summarizes one access; the agent, not the protocol, decides whether to make another.

\begin{algorithm}[t]
\caption{SparseRead Access Protocol}
\label{alg:sparseread-protocol}
{\footnotesize
\noindent\textbf{Input:} artifact \(o\), access intent \(h_a\) (optional), session intent \(h_s\),
agent action \(a\); persistent episode state \(S\)
\begin{algorithmic}[1]
\STATE \(S \gets \operatorname{SyncArtifactVersion}(S,\operatorname{Fingerprint}(o))\)
\STATE \((h,\mathrm{flags}) \gets \operatorname{Validate}(\operatorname{ResolveIntent}(h_a,h_s))\)
\STATE \((H,E,f,L) \gets \operatorname{EnsureEpisode}(S,o,h)\) \COMMENT{fixed full view \(f\)}
\STATE \textbf{if} \(\operatorname{NeedsFullModelView}(h)\) \textbf{or} \(\operatorname{NativeFit}(o,h,L)\) \textbf{then}
\STATE \hspace{\algorithmicindent}\textbf{return} \(\operatorname{FullViewFallback}(f,H)\)
\STATE \textbf{if} \(\operatorname{Underspecified}(h)\) \textbf{or} \(a=\mathrm{scout}\) \textbf{then}
\STATE \hspace{\algorithmicindent}\textbf{return} \(\operatorname{ObjectPreview}(o,\mathrm{flags})\) \COMMENT{bounded and reusable}
\STATE \(r \gets \operatorname{SelectReader}(o,h,a,H)\) \COMMENT{reliability and cost}
\STATE \textbf{if} \(r=\mathrm{none}\) \textbf{then return} \(\operatorname{FullViewFallback}(f,H)\)
\STATE \(b \gets r.\operatorname{Prepare}(o,h,a,H)\)
\STATE \textbf{if} \(b=\mathrm{none}\) \textbf{or} \(\operatorname{EstimatedSize}(b)>L-E\) \textbf{then}
\STATE \hspace{\algorithmicindent}\textbf{return} \(\operatorname{FullViewFallback}(f,H)\)
\STATE \(x \gets \operatorname{SerializeIntact}(b,\mathrm{limit}=L-E)\)
\STATE \textbf{if} \(x=\mathrm{overflow}\) \textbf{then return} \(\operatorname{FullViewFallback}(f,H)\)
\STATE \(\begin{aligned}[t]
H \gets {}&\operatorname{Record}(H,x,h,\operatorname{Fingerprint}(o),\\
&\operatorname{Scope}(b),\operatorname{Coverage}(b,h))
\end{aligned}\)
\STATE \(E \gets E+\operatorname{PayloadSize}(x)\); \(\operatorname{SaveEpisode}(S,H,E)\)
\STATE \textbf{return} \(\begin{aligned}[t]
&\operatorname{EvidencePack}(x,\operatorname{Scope}(b),\\
&\quad\operatorname{Coverage}(H,h),\operatorname{PayloadSize}(x),\mathrm{flags})
\end{aligned}\)
\end{algorithmic}
\noindent\(\operatorname{FullViewFallback}(f,H)\): retire earlier sparse observations;
deliver \(f\) once (or reuse it if already exposed); close episode.
}
\end{algorithm}

\subsection{Adaptive Reader Backends}
\label{sec:reader-backends}

A Reader is defined by the artifacts it handles, not by a particular retrieval algorithm or evaluation scenario. Each Reader registers its supported artifact types, agent actions, and cost profile behind the unified interface. Within a Reader, evidence acquisition has two related choices: \textbf{how to locate or process evidence}, and \textbf{what to expose to the agent}. For example, a text or PDF Reader may locate passages using lexical matching, BM25 windows, or hybrid retrieval. A structured Reader may select columns or key paths, or execute an exact computation; a collection Reader may gather evidence across files. These are location or processing methods within Readers, not separate Reader types.

The exposure choice is distinct. Depending on the intent, a Reader may return anchored snippets, a lossless compact view of the entire artifact, or the result and provenance of an exact operation. Not every method supports every exposure form. We define a \textbf{Reader configuration} \(r\) as a supported combination of a registered Reader, its evidence location or processing method, and an exposure form. All configurations produce the same EvidencePack contract. This separation allows both new artifact types and improved location methods to be introduced without changing the agent's reading interface. The evidence matching differences across artifact types in Section~\ref{sec:reading-challenges} motivate selecting configurations rather than imposing one retrieval pipeline.

SparseRead adapts this choice to the artifact type, size range, and Read Intent. Let \(g\) denote this request group and \(\mathcal{R}_{o,h}\) the compatible configurations for artifact \(o\) and intent \(h\). The reading layer selects

\[ r^* = \arg\min_{r\in\mathcal{R}_{o,h}} \widehat{C}(r \mid g) \quad \mathrm{s.t.} \quad \widehat{p}_{\mathrm{valid}}(r \mid g) \geq \tau, \]

where \(\widehat C\) estimates evidence exposure, Reader execution, and likely follow-up reads under a consistent cost measure. The reliability estimate \(\widehat p_{\mathrm{valid}}\) is calibrated on annotated development requests using an explicit per-request evidence success criterion. Average evidence recall alone is not a probability of valid evidence. When no configuration meets the reliability requirement, the system falls back to a full view rather than silently relaxing the threshold.

Read outcomes can further inform cost estimates: repeated access or escalation indicates extra reading work, whereas a verification request is not itself a failure. Such behavioral signals do not establish evidence correctness. Likewise, source anchors and fingerprints establish provenance, not relevance or sufficiency. This division keeps reliability calibration separate from cost adaptation. Specialized Readers and new evidence methods continue to use the same interface.

\subsection{Budgeted Read Gate}
\label{sec:read-gate}

The Read Gate determines when selective access is worthwhile and when to fall back to a complete view. Let \(f(o)\) be the least expensive \textbf{faithful full view} available for the current request. It may be the original native read or a lossless compact representation of the entire artifact, provided it preserves the access required by the intent. Let \(L=|f(o)|\) be its exposed length. Thus, a compact full view offered by a Reader can lower the Gate's budget. The selected \(f(o)\) and \(L\) remain fixed within an episode; switching to a different full view starts a new episode. Both \(L\) and the cumulative sparse payload \(E\) use the same unit, character length in our implementation, rather than disk size or replay-weighted tokens.

The Gate follows three rules. First, it uses full-view access when the model must inspect the complete content or no suitable Reader is available. Second, it bypasses sparse reading for small artifacts. Otherwise, it begins with sparse evidence and enforces \(E\leq L\). Before delivering a batch, it compares its estimated size with the remaining budget. Serialization then checks the actual payload against the same limit. If the next batch cannot fit intact, it is withheld and the Gate delivers \(f(o)\) instead. Evidence is never truncated in a way that breaks its meaning or source anchors.

Full source processing does not necessarily require full model exposure. A table Reader, for example, can scan every row to calculate a sum and expose only the result, row count, and provenance. The Gate admits this operation to the Reader path because the exposed result is small, even though the processing scope is complete. When a full view has been delivered as evidence, subsequent fallback reuses that view rather than exposing it a second time.

\begin{proposition}[Native-relative exposure bound]
\label{prop:exposure-bound}
Consider a reading episode on a fixed artifact version with a fixed full view \(f(o)\) of length \(L\). If the Gate admits at most \(L\) of cumulative unweighted sparse payload and delivers at most one full view on fallback, the episode's total unweighted artifact exposure is at most \(2L\), excluding separately bounded preview and protocol metadata.
\end{proposition}

The argument is immediate: sparse batches contribute at most \(L\) and a full view adds at most \(L\) (proof in Appendix~\ref{app:exposure-bound}). When an ordinary native full read is among the admissible representations, \(L\) is no greater than its length, so the bound is also at most twice the exposure of that native read. The bound limits the downside of selective reading; savings arise when the agent can finish with much less than \(L\) evidence. It is not a bound on replay-weighted context, billed tokens, task latency, or performance against an oracle that already knows where the answer lies. On fallback, earlier sparse observations are retired from later prompts to avoid continued replay, although their previous cost cannot be recovered. The guarantee applies per episode, not across an entire agent session.

\section{Implementation and Agent Integration}
\label{sec:implementation}

\noindent\textbf{Agent harness integration.} The SparseRead core is a shared
runtime. It owns the Read Intent Protocol, the Read Gate, the evidence record,
the Readers, and the bridge server. Each agent harness adds only the
framework-specific surface needed to route broad object access into that
runtime. NanoBot registers the SparseRead tools directly. OpenCode and OpenClaw
use thin plugin or bridge adapters that expose \texttt{sro\_preview} and
\texttt{sro\_read}, retain \texttt{sro\_card} for compatibility, and translate
local read and open events into the same core calls. Model-facing guidance such
as \texttt{SKILL.md} tells the agent when to start from preview, when to issue
a targeted read, and when to return to native access. This guidance is
advisory: intent validation and the exposure budget are enforced in the core,
so no harness depends on the model's compliance for its guarantees.

\begin{figure*}[t]
  \centering
  \includegraphics[width=\textwidth]{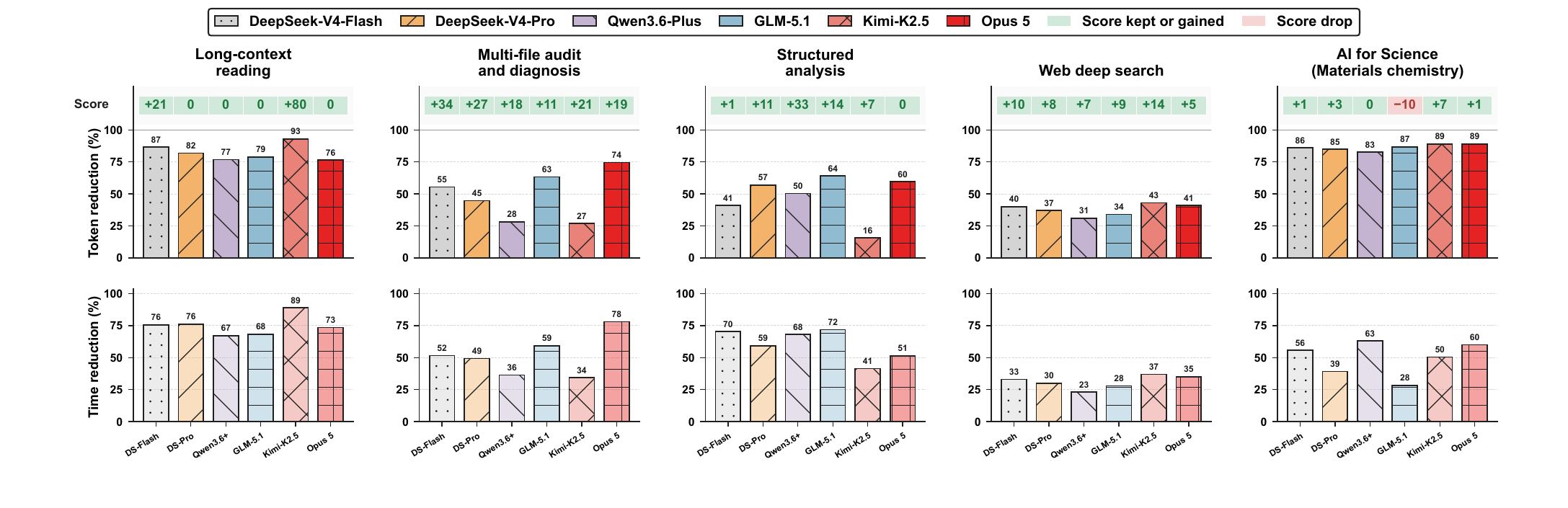}
  \caption{SparseRead results across six models and five scenarios on NanoBot.
  The upper panels report total-token reduction relative to Naive, with markers
  showing the corresponding score delta. The lower panels report end-to-end
  wall-time reduction. Positive bars indicate lower cost.}
  \label{fig:main-results}
\Description{Task scores, token consumption, and wall time across six models and five scenarios.}
\end{figure*}

\noindent\textbf{Readers and the evidence record.} The implementation includes
Readers for text and PDF, structured data (tables and JSON), multi-file
collections, and script libraries. Text and PDF Readers extract page or heading
anchors and return bounded local passages. Table Readers expose the schema,
samples, and row and column projections. They can also run an exact operation
over the full table and return only the result with the row count and source
provenance, or return calculation-ready TSV fragments when the agent computes
itself. Collection Readers build an inventory before selecting candidate files.
Each Reader receives the artifact, the Read Intent, and the current evidence
record, and returns an EvidencePack with source anchors, the scope processed,
and per-slot coverage with unresolved requirements. A Reader declares
its artifact types, actions, and cost profile when it registers, so a new Reader
becomes available to every harness without adapter changes.

\section{Evaluation}

\subsection{Experimental Setup}

The evaluation addresses four research questions. \textbf{RQ1} examines whether SparseRead improves reading efficiency across models and scenarios without sacrificing task quality. \textbf{RQ2} compares SparseRead with retrieval and context compression in both token cost and end-to-end latency. \textbf{RQ3} examines the effects of intent handling, Reader location methods, and Gate budget checks on evidence quality and reading efficiency. \textbf{RQ4} evaluates whether the same SparseRead design transfers across agent frameworks.

\noindent\textbf{Tasks and scenarios.}
We organize the main evaluation around five scenarios, with task sources specified separately for each. \textit{Long-context reading} uses document-comprehension tasks from LooGLE and PinchBench, together with a fact-extraction task adapted from Workspace-Bench-Lite~\cite{li2023loogle,pinchbench2026,tang2026workspacebench}. \textit{Multi-file audit and diagnosis} uses QwenClawBench tasks that require locating and reconciling evidence across workspace files~\cite{qwenclawbench2026}. \textit{Structured analysis} combines QwenClawBench tasks with spreadsheet-manipulation tasks adapted from SpreadsheetBench Verified~\cite{qwenclawbench2026,ma2024spreadsheetbench}. \textit{Web deep search} draws tasks from FRAMES and BrowseComp-Plus, requiring iterative search and evidence synthesis across multiple documents~\cite{krishna-etal-2025-fact,chen2025browsecompplus}. \textit{AI for Science} uses LAMMPS script-generation tasks adapted from MDGYM~\cite{kumar2026mdgym}, with scientific script libraries serving as reading resources. Appendix~\ref{app:task-provenance} gives detailed source and adaptation descriptions. Native-fit controls are treated separately in Appendix~\ref{app:native-fit}, rather than as a main scenario.

\noindent\textbf{Models.}
The main NanoBot experiment includes six models across three groups. Qwen3.6-Plus and DeepSeek-V4-Flash represent \textit{medium models}. DeepSeek-V4-Pro, GLM-5.1, and Kimi-K2.5 represent \textit{leading large open models}, while Claude Opus 5 represents a \textit{leading commercial closed model}. This selection tests whether sparse reading remains effective across model sizes and whether its benefits persist for leading models with strong reasoning and tool use capabilities.

\noindent\textbf{Agent frameworks.}
We integrate SparseRead into \textit{NanoBot v0.2.0, OpenCode v1.17.14,} and \textit{OpenClaw v2026.6.11}. All integrations share the same reading protocol and Reader Backends, while thin adapters connect the protocol to each framework's tool interface. The cross-framework experiment uses six models and requires no model retraining.

\noindent\textbf{Baselines.}
We compare SparseRead with Naive and four alternatives. \textit{Naive} preserves the unmodified framework's agent behavior, including autonomous search and partial reads through native tools. \textit{RAG Static} retrieves and reranks evidence once, then generates from a fixed evidence budget without further source access. \textit{RAG Agent} retrieves iteratively with the same retrieval service, retaining native tools under SparseRead's iteration and timeout limits. \textit{ACON} compresses both environment observations and interaction history during agent execution~\cite{kang2026acon}. \textit{Obs. masking} follows the fixed positional policy from \textit{Complexity Trap}, retaining the latest ten observations while masking older ones~\cite{lindenbauer2025complexity}. Configuration details appear in Appendix~\ref{app:baseline-config}.

\noindent\textbf{Metrics.}
All costs are measured end to end, including retrieval and reranking for the RAG baselines and auxiliary model calls for ACON. We report total token reduction, task-score change, and wall-time saving relative to Naive; a positive delta indicates an improvement. Figure score strips express score changes in percentage points. Component ablations additionally examine protocol diagnostics, evidence hit rate, and raw artifact exposure.

\begin{figure*}[t]
  \centering
  \includegraphics[width=\textwidth]{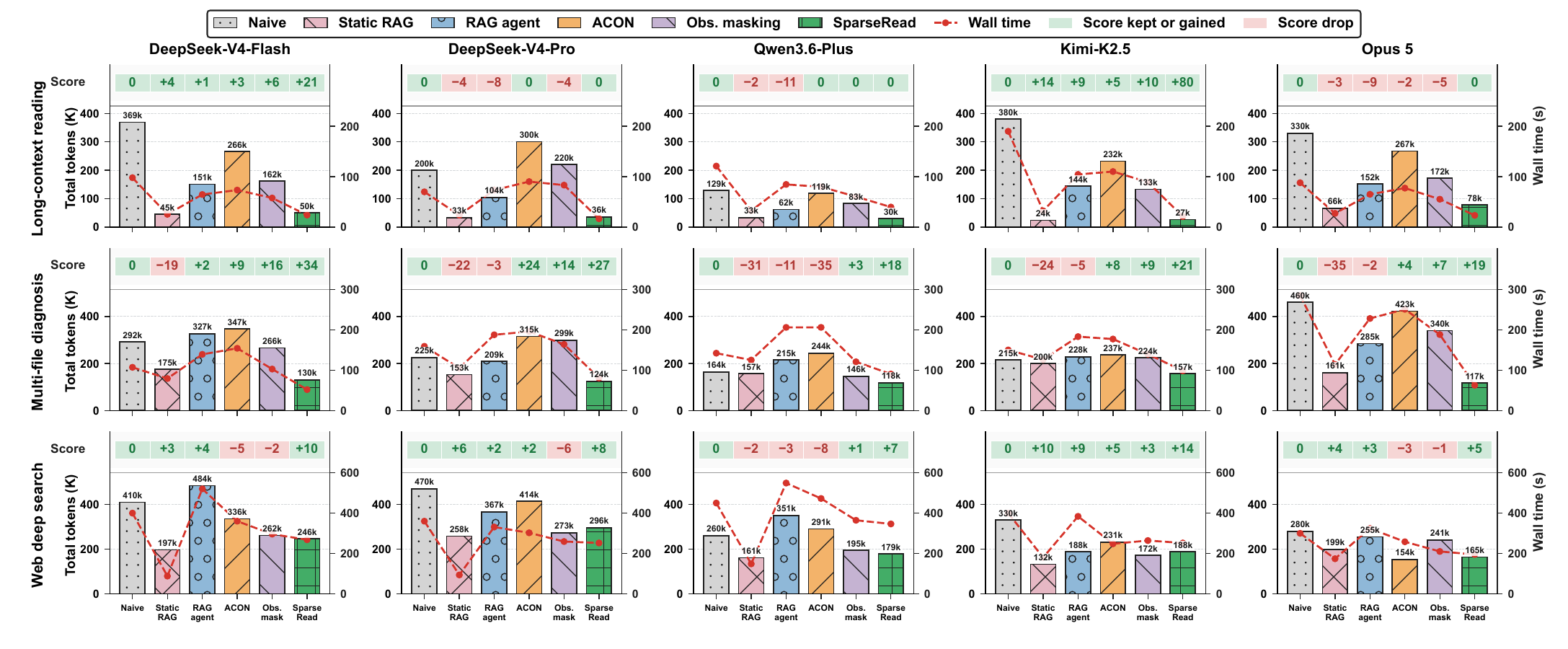}
  \caption{Six methods are compared across three scenarios and five models by total tokens, wall time, and score changes.}
  \label{fig:baseline-comparison}
\Description{Comparison with ACON and CT context-reduction baselines across two scenarios and three models.}
\end{figure*}

\begin{table}[t]
  \centering
  \caption{End-to-end SparseRead results across three agent frameworks and six models.}
  \label{tab:cross-framework}
  \footnotesize
  \begin{tabular*}{\columnwidth}{@{\extracolsep{\fill}}llccc@{}}
  \toprule
  Framework & Model & Token red. & Score $\Delta$ & Time save \\
  \midrule
  \textbf{NanoBot} & Qwen3.6-Plus & 69.0\% & +0.039 & 64.4\% \\
   & DeepSeek-Flash & 22.4\% & +0.140 & 25.6\% \\
   & DeepSeek-Pro & 71.3\% & $-0.015$ & 73.1\% \\
   & GLM-5.1 & 61.0\% & +0.012 & 52.6\% \\
   & Kimi-K2.5 & 92.9\% & +0.598 & 85.4\% \\
   & Opus 5 & 58.6\% & +0.024 & 54.2\% \\
  \addlinespace[2pt]
  \textbf{OpenCode} & Qwen3.6-Plus & 25.2\% & +0.333 & 22.8\% \\
   & DeepSeek-Flash & 71.8\% & +0.143 & 64.9\% \\
   & DeepSeek-Pro & 58.5\% & $-0.013$ & 62.9\% \\
   & GLM-5.1 & 75.4\% & +0.033 & 68.9\% \\
   & Kimi-K2.5 & 83.7\% & $-0.031$ & 73.7\% \\
   & Opus 5 & 46.8\% & +0.018 & 42.7\% \\
  \addlinespace[2pt]
  \textbf{OpenClaw} & Qwen3.6-Plus & 46.7\% & +0.032 & 56.8\% \\
   & DeepSeek-Flash & 28.7\% & +0.062 & 26.5\% \\
   & DeepSeek-Pro & 14.5\% & +0.151 & 28.2\% \\
   & GLM-5.1 & 28.7\% & +0.118 & 38.6\% \\
   & Kimi-K2.5 & 42.8\% & +0.032 & 24.6\% \\
   & Opus 5 & 32.5\% & +0.011 & 29.4\% \\
  \bottomrule
  \end{tabular*}
\end{table}

\subsection{Effectiveness Across Models and Scenarios}

Figure~\ref{fig:main-results} covers 30 settings across six models and five scenarios. SparseRead reduces tokens by 15.7--92.9\% and wall time by 23.0--89.0\%, with median reductions of 61.4\% and 53.7\%. Score is preserved or improved in 29 settings; regressions occur for GLM-5.1 on AI for Science.

Long-context tasks save 76.5--92.9\% of tokens and 67.1--89.0\% of time. Multi-file diagnosis and structured analysis preserve or improve score for every model, with token reductions of 26.8--74.5\% and 15.7--64.1\%, respectively.

On Web deep search, all six models gain 5--14 percentage points while saving 31.0--43.0\% of tokens and 23.0--37.0\% of time. AI for Science saves 82.7--89.1\% of tokens and 28.3--63.0\% of time, with score preserved or improved for five models. Opus 5 saves 41.0--89.0\% of tokens and 35.0--78.1\% of time across all five scenarios, showing that stronger models still benefit from reading less.

\subsection{Comparison with Retrieval and Context-Reduction Baselines}

Figure~\ref{fig:baseline-comparison} compares six methods in 15 settings across three scenarios and five models. SparseRead preserves or improves score in every setting, saving 27.0--92.9\% of tokens and 23.0--88.9\% of time. It uses the fewest tokens in six settings and the least time in nine, achieving both minima for all five models on multi-file diagnosis with score gains of 17.8--34.5 percentage points.

RAG Static is faster in all five Web settings but scores below Naive in nine of 15 settings. Under the same iteration and timeout limits, RAG Agent loses score in eight settings and exceeds Naive's time in eight. ACON loses score in five settings and increases both costs in six; observation masking loses score in five. SparseRead's advantage is consistent with controlling admission through explicit evidence needs and coverage, beyond iterative retrieval or compression after acquisition.

\begin{figure*}[t]
  \centering
  \includegraphics[width=0.9\textwidth]{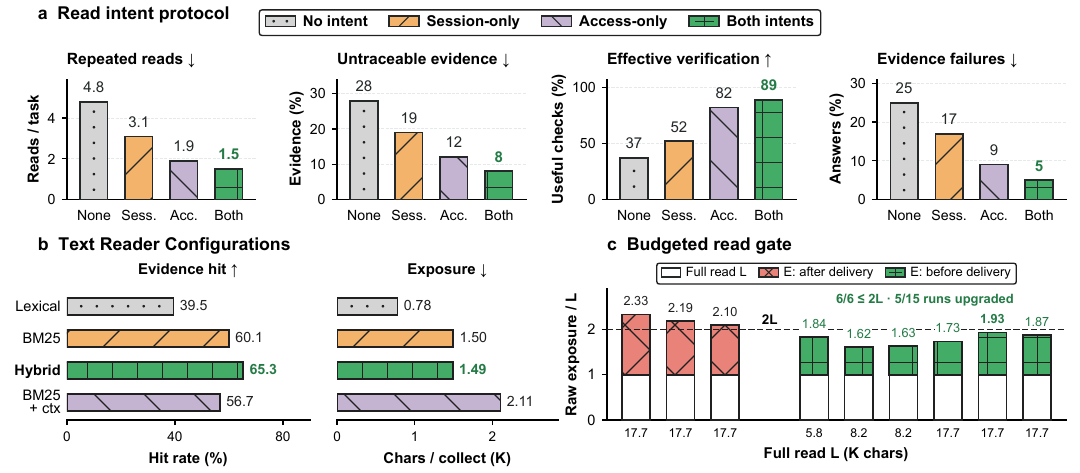}
  \caption{Ablations of intent handling (a), Reader location methods (b), and Gate budget checks (c). Panel (c) compares cumulative raw exposure against the \(2L\) bound for checks before and after evidence delivery.}
  \label{fig:ablations}
\Description{Native and gate-selected SparseRead costs and scores across eight matched tasks.}
\end{figure*}

\subsection{Portability Across Agent Frameworks}

Table~\ref{tab:cross-framework} shows lower token use and wall time in all 18 settings, with score preserved or improved in 15. For NanoBot, OpenCode, and OpenClaw, median token reductions are 65.0\%, 65.2\%, and 30.6\%; time savings are 59.3\%, 63.9\%, and 28.8\%. Three score decreases range from 0.013 to 0.031. Opus 5 saves 32.5--58.6\% of tokens and 29.4--54.2\% of time, with score gains of 0.011--0.024. Reusing the protocol, Readers, and Gate through thin adapters supports portability without model retraining.

\subsection{Ablation of the Three Core Designs}
\label{sec:ablations}

\noindent\textbf{Read Intent Protocol.}
Figure~\ref{fig:ablations}(a) compares no intent, session intent, access intent, and both. Combining intents reduces repeated reads from 4.8 to 1.5 per task (68.8\%) and untraceable evidence from 28\% to 8\%. The share of verification checks that resolve an evidence gap rises from 37\% to 89\%, while answers failing from missing evidence fall from 25\% to 5\%. Access intent improves every diagnostic more than session intent alone. Adding session intent further lowers repeated reads from 1.9 to 1.5 and raises effective verification from 82\% to 89\%, giving the best result on all four diagnostics. This supports combining the broader session objective with the current evidence need to reduce redundant access and improve verification.

\noindent\textbf{Adaptive Reader Backends.}
Figure~\ref{fig:ablations}(b) compares four text Reader location methods on 47 LooGLE documents and 741 questions. A hit requires at least 80\% annotated evidence coverage; exposure is median returned characters per collect over five questions. Lexical matching achieves 39.5\% hit rate at 780 characters, BM25 60.1\% at 1,497, and Hybrid 65.3\% at 1,490. Hybrid thus gains 5.2 percentage points over BM25 at essentially the same exposure. Expanding BM25's windows instead increases exposure to 2,106 characters while lowering hit rate to 56.7\%. This comparison quantifies location method quality. It does not isolate the incremental benefit of the configuration selector in Section 4.2 over fixed configurations.

\noindent\textbf{Budgeted Read Gate.}
Figure~\ref{fig:ablations}(c) compares checks after and before delivery in separate cohorts, each using five tasks and three repeats on DeepSeek-V4-Flash. Bars normalize cumulative raw exposure by the full view as \((E+L)/L\), using character counts. Checking after delivery allows overshoot: three upgrades reach 2.10--2.33 times the full view. Checking before delivery keeps all six exposure ratios within 1.62--1.93, with upgrades in five of 15 runs.

For tasks unsuited to sparse reading, the Gate bypasses the sparse path and retains native tool use to avoid unnecessary protocol overhead; Appendix~\ref{app:native-fit} reports the corresponding token-cost and task-score comparison with forced sparse reading.

\section{Related Work}

\noindent\textbf{Prompt and KV compression.}
Selective Context, LLMLingua, Gist Tokens, AutoCompressors, and ICAE reduce prompt text or learn compact representations~\cite{li-etal-2023-compressing,jiang-etal-2023-llmlingua,jiang-etal-2024-longllmlingua,pan-etal-2024-llmlingua,NEURIPS2023_3d77c6dc,chevalier-etal-2023-adapting,ICLR2024_0b276510}; KV methods optimize cached attention states~\cite{NEURIPS2023_6ceefa7b,ICLR2024_5e5fd18f,NEURIPS2024_28ab4182,cai2024pyramidkv,pmlr-v235-liu24bz,NEURIPS2024_7e57131f,liu2024cachegen,liu2026kvserve,wu2026beyond}. SparseRead governs external artifact access and complements these inference optimizations. Ordinary sparse accesses preserve preceding turns, enabling prefix reuse~\cite{MLSYS2024_a66caa17,liu2026elasticmm}.

\noindent\textbf{Agent observations and histories.}
SWE-Pruner and Squeez prune tool outputs according to task needs~\cite{wang2026swepruner,kovacs2026squeez}; SWE-Pruner Pro selects lines using the agent's internal representations~\cite{wang2026sweprunerpro}. CoACT, AgentDiet, TACO, ACON, and AgentFold reduce observations or histories~\cite{chen2026coact,xiao2026reducing,ren2026selfevolving,kang2026acon,ye2025agentfold}, while observation masking drops older observations~\cite{lindenbauer2025complexity}. SparseRead provides a reading control layer above individual compressors: its \emph{pluggable Reader interface} lets SWE-Pruner and Squeez serve as compression backends. The \emph{Read Intent Protocol} coordinates these backends through versioned evidence records linking intent, source coverage, and unresolved needs across refinement and verification.

\noindent\textbf{Active reading and retrieval.}
ReadAgent combines gist memories with selective lookup~\cite{lee2024readagent}; IRCoT and Self-RAG interleave retrieval with reasoning or reflection~\cite{trivedi2023ircot,asai2024self}; Adaptive-RAG routes retrieval by question complexity~\cite{jeong2024adaptiverag}. Retrieval is one acquisition method under SparseRead's unified evidence contract, alongside structured projections and exact source operations. \emph{Processing scope and model exposure are controlled separately}: a Reader may scan an entire table yet expose only the result and provenance. Its \emph{Budgeted Read Gate} adapts classical ski rental budgeting~\cite{borodin1998online} to cumulative raw exposure across reads, checking each batch before delivery and retaining a faithful full view as fallback within \(2L\) per episode.

\section{Conclusion}

SparseRead provides a reusable reading control layer between agents and external artifacts. Its stateful protocol, pluggable Readers, and Budgeted Read Gate coordinate evolving evidence needs, separate source processing from model exposure, and bound raw exposure while retaining a faithful full view as fallback. The unified interface lets retrieval and compression methods serve as replaceable backends, so advances in evidence acquisition can strengthen the framework across agent implementations. Evaluations show token and latency savings even for leading models, with task quality preserved or improved in most settings; ablations support the complementary roles of the three designs.

\bibliographystyle{ACM-Reference-Format}
\bibliography{references}

\appendix
\section{Proof of the Native-Relative Exposure Bound}
\label{app:exposure-bound}

\begin{figure*}[t]
  \centering
  \includegraphics[width=0.8\textwidth]{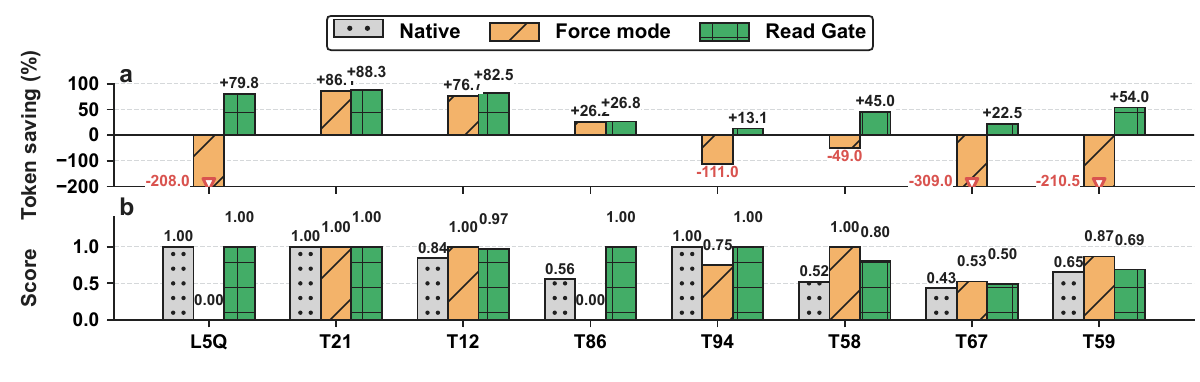}
  \caption{Read-Gate behavior across 8 matched task on Deepseek-v4 Pro.  Gray bars
  denote Native, and colored bars denote SparseRead under the gate-selected
  policy. Annotations report score changes or cost reductions relative to
  Naive.}
  \label{fig:gate-behavior}
\end{figure*}

\begin{proof}[Proof of Proposition~\ref{prop:exposure-bound}]
Fix an episode, so that $f(o)$ and $L$ are fixed. Exposure is measured in the unit of $L$ (characters), and background processing by a Reader is not exposure. As in the proposition, separately bounded preview and protocol metadata are excluded. We first show that $E \le L$ at all times. Initially $E=0$. The counter changes only at line 16 of Algorithm~1, after line 13 has serialized the batch under the limit $L-E$. The updated value $E+\mathrm{PayloadSize}(x)$ is therefore at most $L$. Every other branch, including each fallback, leaves $E$ unchanged. Now count what the episode exposes. Admitted sparse evidence accounts for exactly $E \le L$. The fallback delivers $f(o)$ at most once and closes the episode. If $f(o)$ was already exposed as admitted evidence, the fallback adds nothing and the total is $E \le L$. Otherwise it adds $L$, and the total is at most $E + L \le 2L$. An episode that ends without a fallback exposes $E \le L$. In every case the total is at most $2L$.
\end{proof}

\noindent The invariant depends only on the check of the serialized payload (lines 13 and 14). The size estimate at line 11 can only trigger a fallback earlier. Delivering a batch that crosses the budget and escalating afterwards would instead give $2L+\delta$, where $\delta$ is the largest batch size. When an ordinary native read is an admissible full view, $L$ is at most its length $N$, so the exposure is at most $2N$. The constant 2 is tight for this rule: admitted batches totaling just under $L$, followed by a withheld batch and a fallback, expose just under $2L$. The count is cumulative and unweighted. Retiring earlier observations on fallback reduces future replay but does not refund exposure already paid. The bound therefore says nothing about replay-weighted context, billed tokens, latency, or sessions with several episodes.

\section{Task Sources and Scenario Construction}
\label{app:task-provenance}

\noindent\textbf{Coverage and inclusion.}
For each scenario, we include \emph{all relevant tasks} from every benchmark listed below, giving \emph{100\% coverage of its scenario-relevant task set}. Relevance is determined by the input artifacts and required operations described below, rather than by task scores or observed reading savings; tasks are not sampled or chosen as favorable examples. The coverage denominator is the relevant task set within each benchmark, not unrelated task families in the complete benchmark. Adaptations map the scenario-relevant requirements, source artifacts, and outputs to the agent interface and checker. Multiple questions or follow-up variants over the same source remain related task formulations, rather than independent source documents.

\noindent\textbf{Long-context reading.}
We include all long-document comprehension tasks from LooGLE~\cite{li2023loogle}, all document-comprehension tasks from PinchBench~\cite{pinchbench2026}, and all annual-report fact-extraction tasks relevant to this scenario from Workspace-Bench-Lite~\cite{tang2026workspacebench}: 100\% of the relevant tasks in each source. LooGLE includes single-question, multi-question, and follow-up formulations. Workspace-Bench-Lite tasks are adapted to extracting requested facts, such as research-and-development figures, from company annual reports; they do not reproduce unrelated operations from the complete workspace benchmark.

\noindent\textbf{Multi-file audit and diagnosis.}
We include all QwenClawBench tasks requiring audit or diagnosis across workspace files~\cite{qwenclawbench2026}, covering 100\% of its tasks relevant to this scenario. These comprise data-integrity checks, configuration and scheduled-job failures, command-prefix security analysis, and service-audit diagnosis. Each requires locating and reconciling evidence across files, rather than merely retrieving an isolated fact.

\noindent\textbf{Structured analysis.}
We include all relevant statistical and financial analysis tasks from QwenClawBench and all spreadsheet-manipulation tasks from SpreadsheetBench Verified that operate on the structured artifacts in this scenario~\cite{qwenclawbench2026,ma2024spreadsheetbench}, covering 100\% of the relevant tasks in each source. Spreadsheet tasks, including lookup and weekday-row correction, are adapted to the agent's execution interface and checker. Coverage refers to these structured tasks, rather than unrelated capabilities in either benchmark.

\noindent\textbf{Web deep search.}
We include all relevant multi-source search and reasoning tasks from FRAMES and BrowseComp-Plus, covering 100\% of the relevant tasks in each benchmark. FRAMES supplies multi-hop questions requiring evidence from multiple sources~\cite{krishna-etal-2025-fact}; BrowseComp-Plus supplies deep-research questions with human-verified supporting documents and challenging negatives in a fixed corpus~\cite{chen2025browsecompplus}. Together they evaluate iterative search and selective reading across documents. BrowseComp-Plus uses its fixed corpus rather than unrestricted live-Web browsing.

\noindent\textbf{AI for Science.}
We include all tasks in MDGYM's LAMMPS portion relevant to script generation, covering 100\% of this relevant task set~\cite{kumar2026mdgym}; GROMACS tasks fall outside the scenario. Scientific script libraries provide the external reading resources. The adaptation evaluates script generation and reading behavior, rather than reproducing MDGYM's complete simulation-execution and physical-validation evaluation.

\section{Native-Fit Controls and Gate Economics}
\label{app:native-fit}
Native-fit controls are drawn from QwenClawBench~\cite{qwenclawbench2026} and include finding the largest file in a directory, implementing a user-discount calculator, and writing a SPARQL query. They probe settings in which direct tool use or a short native read can already be appropriate. They are excluded from the five main scenarios. In the original five-scenario matrix, including these controls, SparseRead reduces token volume and wall time in all 30 model--scenario cells and preserves or improves score in 26. The corresponding results for the four retained scenarios, excluding Web deep search, preserve or improve score in 22 of 24 cells.

\noindent\textbf{Economic boundary of selective reading.}
The Gate checks whether sparse access is worthwhile before invoking a Reader. When native tools or a short native read already suit the request, it retains that path instead of forcing preview, retrieval, or refinement. This bypass aims to avoid the protocol tax of sparse reading; it does not require the native agent to read every file in full.

Figure~\ref{fig:gate-behavior} reports the supplementary eight-task NanoBot comparison on DeepSeek-V4-Flash, using Naive, forced sparse reading without admission gating, and the earlier Gate policy. The tasks are LooGLE five-question QA (L5Q), PDF comprehension (T21), stock-system audit (T12), command security (T86), exam-system audit (T94), DiD regression (T58), SPARQL generation (T67), and discount calculation (T59). The last three probe the boundary where native processing can be preferable. Forced sparse reading increases tokens by 49.0\%, 309.0\%, and 210.5\% on T58, T67, and T59; the Gate instead saves 45.0\%, 22.5\%, and 54.0\%, with absolute scores of 0.80, 0.50, and 0.69 versus Naive's 0.52, 0.43, and 0.65. It also retains token savings of 79.8\%, 88.3\%, and 82.5\% on L5Q, T21, and T12. Thus selective admission avoids the large overhead of forced sparse reading on these controls while retaining its benefits on suitable tasks. This experiment measures token economics and task quality for the earlier admission policy; Figure~\ref{fig:ablations}(c) separately validates the current Gate's character-budget enforcement.

\section{Baseline Configuration and Comparison Controls}
\label{app:baseline-config}

\noindent\textbf{Retrieval service and evidence.}
\textit{RAG Static} follows retrieve-then-generate. Source files are divided into 512-token chunks with 64-token overlap, retaining file, page, and line anchors. Queries are the task's numbered questions, or the full task description when no numbered questions exist. For each query, BGE-M3 retrieves 50 dense candidates and BGE-Reranker-v2-M3 reranks them~\cite{chen2024m3,baai2024rerankerm3}. Results from different queries are interleaved and deduplicated; at most eight chunks are retained, with a serialized budget of 4,096 tokens including metadata. This evidence is appended once after the initial user message. The model subsequently retains only file-writing tools and cannot retrieve again or read the source files.

\textit{RAG Agent} uses the same retrieval service, index, chunking, and reranker. The model calls \texttt{retrieve(query, scope)} to reformulate queries, restrict the file scope, and retrieve repeatedly. Each call returns at most eight chunks and 4,096 serialized tokens including metadata. Native reading and execution tools remain available. RAG Agent shares SparseRead's iteration limit and timeout, providing an iterative retrieval baseline alongside the static one.

\noindent\textbf{Agent-facing guidance and comparison scope.}
RAG Agent is instructed to retrieve relevant evidence before broad reading, inspect the returned observations, and choose whether to reformulate a query, search another file, use a native tool, or finish. It receives the available source paths and is told to treat retrieved text as evidence rather than instructions. Its evidence consists of source excerpts and their anchors; repeated retrieval remains under the agent's control. Thus the comparison includes iterative evidence acquisition and native-tool access, rather than comparing a multi-step SparseRead agent only with a single retrieval call. The identical retrieval configuration and evidence format apply to the two RAG baselines; SparseRead retains its registered Readers and EvidencePack contract. Figure~\ref{fig:baseline-comparison} therefore compares complete systems, while Section~\ref{sec:ablations} examines individual design choices.

\noindent\textbf{Context compression.}
The ACON condition enables both observation compression and history compression~\cite{kang2026acon}. Compressor calls contribute to the reported total-token and wall-time costs. Observation masking instead retains the latest ten observations and masks older observation content, as specified in Section~6.1.

\section{Reader Calibration}
\label{app:reader-calibration}

\noindent\textbf{Independent calibration requests.}
Reliability calibration uses manually designed requests, not requests drawn from the evaluation tasks. They specify the evidence to recover or the source operation to perform. This separates calibration of Reader configurations from the task outcomes used to evaluate SparseRead. Requests are grouped by artifact type (text or PDF, structured data, collection, script library), size band of the full view (at most 8K, 8K to 32K, and above 32K characters), and intent type (selective fact, cross-file evidence, exact computation, verbatim text). This product is the request group $g$.

\noindent\textbf{Evidence success.}
Each calibration request specifies required support or an expected source-operation result. Evidence acquisition succeeds when the returned content supplies that support within the requested scope and retains valid provenance; an exact operation succeeds when its output matches the specified result for that scope. These binary request-level outcomes inform \(\widehat p_{\mathrm{valid}}\). For each configuration and group, \(\widehat p_{\mathrm{valid}}\) is the lower end of a 95\% Wilson interval over these outcomes, and the threshold is \(\tau=0.6\), fixed before evaluation. A group with fewer than 20 calibration requests is not adaptive: it uses the default configuration for its artifact type, and falls back to a full view if that configuration does not meet the threshold. Cost profiles account for evidence exposure and further Reader work, while evaluation reports total tokens and wall time separately. The estimate \(\widehat C\) is measured in characters: exposed characters, a fixed per-call overhead for Reader execution, and the mean follow-up exposure observed for the group during calibration. Reliability is calibrated once; evaluation updates only cost estimates.

The location-method experiment in Figure~\ref{fig:ablations}(b) is a separate offline evaluation: its 80\% gold-evidence coverage criterion defines an evidence hit, not the selector's reliability threshold \(\tau\). Its results compare location configurations rather than measuring the incremental gain from adaptive selection.

\end{document}